\documentclass[final]{cvpr2021}

\usepackage{times}
\usepackage{epsfig}
\usepackage{graphicx}
\usepackage{amsmath}
\usepackage{amssymb}
\usepackage{color}

\usepackage{algorithm}
\usepackage{algorithmic}

\usepackage{multirow}

\usepackage[pagebackref=true,breaklinks=true,colorlinks,bookmarks=false]{hyperref}

\newcommand{\nj}[1]{\textcolor{green}{#1}}

\newcommand{\quotes}[1]{``#1''}

\def\cvprPaperID{7467} 
\def\confYear{CVPR 2021}

\begin{document}

\title{Interpolation-based semi-supervised learning 
for object detection}

\author{
Jisoo Jeong\\
Seoul National University\\
{\tt\small soo3553@snu.ac.kr}
\and
Vikas Verma\\
Aalto University, Finland, \\
Mila, Université de Montréal \\
{\tt\small vikas.verma@aalto.fi}
\and
Minsung Hyun\\
Seoul National University, \\
SK hynix\\
{\tt\small minsung.hyun@snu.ac.kr}
\and
Juho Kannala \\
Aalto Univeristy, Finland \\
{\tt\small juho.kannala@aalto.fi}
\and
Nojun Kwak \thanks{corresponding author}\\
Seoul National University\\
{\tt\small nojunk@snu.ac.kr}
}

\maketitle

\begin{abstract}
Despite the data labeling cost for the object detection tasks being substantially more than that of the classification tasks, semi-supervised learning methods for object detection have not been studied much. 
In this paper, we propose an Interpolation-based Semi-supervised learning method for object Detection (ISD), which considers and solves the problems caused by applying conventional Interpolation Regularization (IR) directly to object detection. 
We divide the output of the model into two types according to the objectness scores of both original patches that are mixed in IR. 
Then, we apply a separate loss suitable for each type in an unsupervised manner. 
The proposed losses dramatically improve the performance of semi-supervised learning as well as supervised learning. 
In the supervised learning setting, our method improves the baseline methods by a significant margin.
In the semi-supervised learning setting, our algorithm improves the performance on a benchmark dataset (PASCAL VOC and MSCOCO) in a benchmark architecture (SSD).


\end{abstract}

\section{Introduction}
\label{sec:intro}
A dataset for object detection is much harder to create than the one for classification. 
While there is only one class in a single image for the classification task, there are multiple objects with different class labels in a single image for the object detection task.
Therefore, the dataset for supervised object detection requires a pair of a class label and bounding box information for each object.
Labeling each object takes more than a few seconds, and creating these datasets requires hundreds of hours \cite{russakovsky2015best, bearman2016s, dollar2012pedestrian}.

\begin{figure}[t]
\begin{center}
\includegraphics[width=0.8\linewidth]{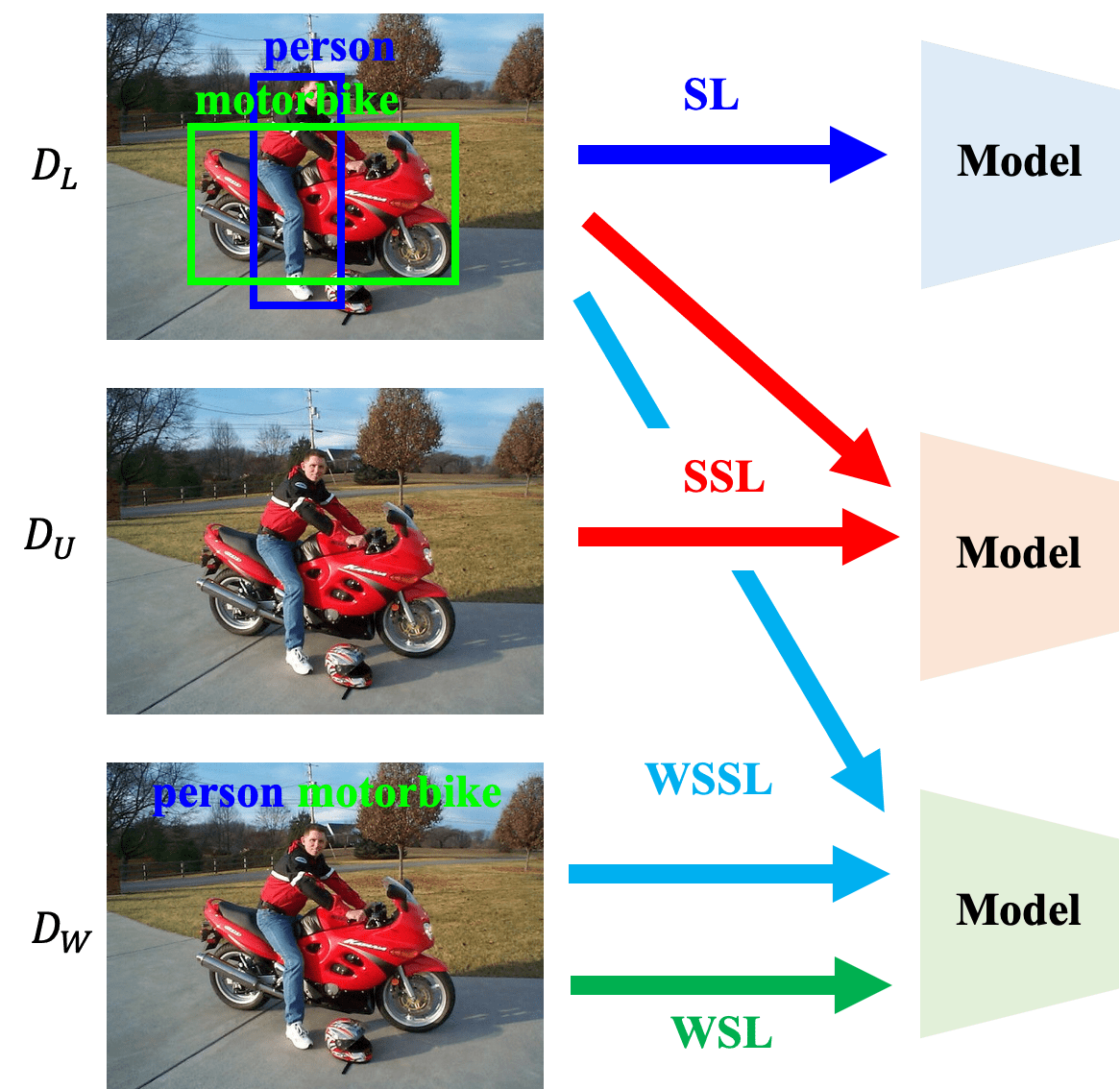}
\end{center}
\caption{Supervised Learning (SL), Semi-supervised Learning (SSL), Weakly-Supervised Learning (WSL) and Weakly Semi-Supervised Learning (WSSL) for Object Detection. In this paper, we deal with SSL.}

\label{intro:dataset}
\end{figure}

\begin{figure*}[ht]
\begin{center}
\includegraphics[width=0.82\linewidth]{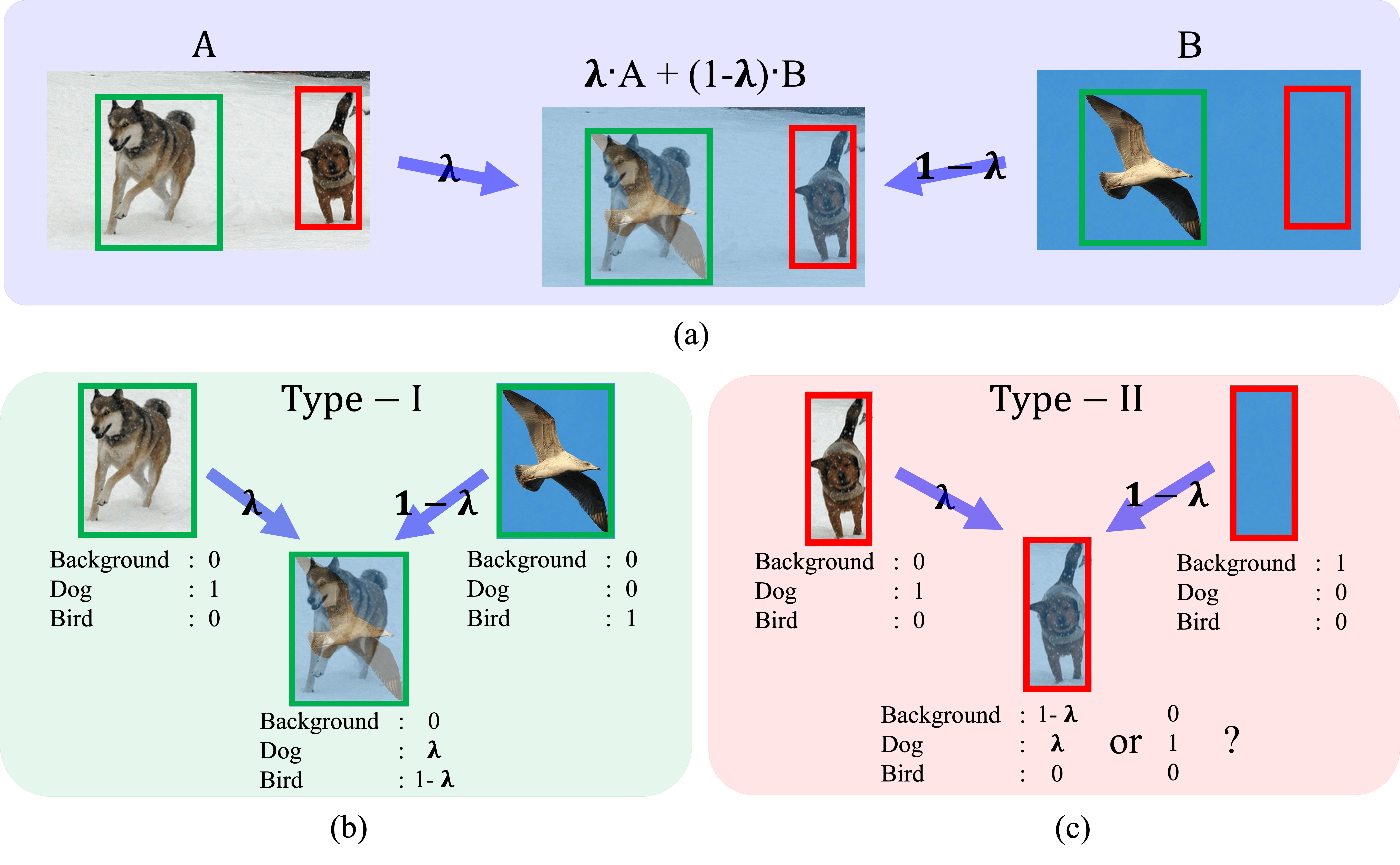}
\end{center}
\caption{(a) Mixed image created by random interpolation between images A and B (b) Type-I : both patches are from object classes. (c) Type-II : one of the patches is from the background class.
}
\label{intro:problem}
\end{figure*}

Due to the higher time and resource complexity for creating object detection datasets, recently, methods for learning with weakly labeled data ($D_{W}$) or unlabeled data ($D_{U}$) have been studied as opposed to learning with the labeled data ($D_L$)\footnote{$D_{L} = {(I_{i}, y_{i})}_{i=1}^{N_{L}}$ where $y_{i} = {(class^{j}, bbox^{j})}_{j=1}^{M_i}$ , $D_{W} = {(I_{i}, y_{i})}_{i=1}^{N_{W}}$ where $y_{i} = {(class^{j})}_{j=1}^{M_i}$, and $D_{U} = {(I_{i})}_{i=1}^{N_{U}}$. Here, $N_X$ is the number of images, and $M_i$ is the number of objects in the image $I_i$.} only. 
There are mainly three types of object detection methods: weakly-supervised, semi-supervised, and weakly-semi-supervised learning.
Weakly-supervised learning trains a model with a dataset that has only class information but no location information ($D_{W}$) \cite{zhu2017soft, shi2017weakly, jie2017deep, wang2018collaborative, kim2019tell}.
On the other hand, weakly-semi-supervised learning is a learning method which uses $D_{W}$ as well as $D_L$ \cite{tang2016large, yan2017weakly}.
Weakly-semi-supervised detector improves its performance compared to that of weakly-supervised learning, but it still needs to label classes for $D_W$.
In the setting of semi-supervised object detection, instead of $D_W$, unlabeled data $D_U$ is utilized in combination with the labeled data ($D_{L}$) \cite{wang2018towards, Nguyen2019em, jeong2019consistency} (See Fig. \ref{intro:dataset}.).

In this paper, we address the semi-supervised object detection problem and propose a new method called Interpolation-based Semi-supervised learning for object Detection (ISD) whose loss terms can also be applied to the supervised learning framework.
Interpolation Regularization (IR) which mixes different images and learns to predict the combined label rather than one hot vector performs outstandingly in supervised learning as well as in semi-supervised learning for classification problems \cite{zhang2017mixup, verma2018manifold, verma2019interpolation, berthelot2019mixmatch, Verma2019GraphMixRT}.
However, it is challenging to apply IR directly to object detection because the background class consists of a very diverse and irregular texture.
Fig.~\ref{intro:problem} shows an example of applying IR to the object detection problem.
In Fig.~\ref{intro:problem}(a), we mix image $A$ and $B$ using the mixing parameter $\lambda = 0.5$ as shown in the middle. Obviously, the mixed green box has 50\% of a dog and 50\% of a bird as we can see in Fig.~\ref{intro:problem}(b). 
However, when an object is mixed with a background as in Fig.~\ref{intro:problem}(c), the mixed image appears to be an 100\% object corrupted by noise.
If the detector is trained by the conventional IR, any blurred or noisy mixture images contribute to increasing the confidence of the background class, and it will degrade performance.
On the other hand, if that sample is trained as a foreground object, it is expected to be robust to noise and to learn about various backgrounds around the object.

To tackle this problem, in this paper, we divide the mixed images into two types (Type-I and II) depending on whether one of the original images is the background or not.
Then, we apply a different IR algorithm suitable for each type.
The proposed ISD method which will be detailed in Sec.~\ref{sec:method} can be combined with conventional semi-supervised learning methods such as CSD (consistency-based semi-supervised learning) \cite{jeong2019consistency} to improve the semi-supervised object detection performances.
Also, the proposed scheme can be used to enhance the detection performance in the supervised learning framework.
Our main contributions can be summarized as follows:




\begin{itemize}
\item We show the problem in applying interpolation regularization directly to the object detection task and propose a novel interpolation-based semi-supervised learning algorithm for object detection.

\item In doing so, we define two types of interpolation in the object detection task and propose efficient semi-supervised learning methods suitable for each type.

\item We experimentally show the effectiveness of the proposed method for each type by demonstrating a significant performance improvement over the conventional algorithms achieving SOTA semi-supervised object detection performance.
\end{itemize}

\section{Related Work}
\label{sec:related}

\subsection{Interpolation-based Regularization (IR)}
Interpolation-based Regularization is a promising approach due to its state-of-the-art performances and virtually no additional computational cost. These methods construct additional training samples by combining two or more training samples. Mixup \cite{zhang2017mixup} and Between-class learning \cite{tokozume2017betweenclass} are the earliest works that took steps in this direction. These methods are based on the principle that the output of a supervised network for an affine combination of two training samples should change linearly. 
Such kind of inductive bias can be induced in a network by training it on the synthetic samples constructed by \textit{mixing} two samples and their corresponding targets. Manifold Mixup \cite{verma2018manifold} mixes features in the deeper layers instead of input images. 
Other works such as CutMix \cite{cutmix} construct the synthetic samples by mixing the CutOut \cite{devries2017cutout} versions of two samples. Overall, these approaches can be interpreted as a form of data-augmentation technique that seeks to construct additional training samples by combining two or more samples. In the unsupervised learning setting, interpolation-based regularizers have been explored in ACAI \cite{acai} and AMR \cite{beckham2019adversarial}. These methods learn better unsupervised representations by enforcing a constraint that the representations obtained by mixing the representations of two samples should correspond to a data point on the data manifold.

\subsection{Semi-Supervised Learning (SSL)}
Semi-Supervised Learning (SSL) is a dominant approach for machine learning when the annotated data is scarce. There has been recent surge of interest in deep learning based on SSL for object classification \cite{verma2019interpolation, berthelot2019mixmatch, Verma2019GraphMixRT}. These methods can be broadly categorized into: (1) consistency regularization methods \nj{and} (2) generative adversarial networks (GAN) based methods. 
We describe below focusing on consistency regularization methods, which is highly relevant to out research.

The central idea of the consistency regularization methods is to enforce that the model predictions should not change under \textit{reasonable} permutations to the input. For object classification, such permutations entail random translation, random cropping and horizontal flipping etc. Let us assume that $x$ and $x'$ are the original and the permuted inputs, $d(\cdot,\cdot)$ be a distance function, $w(t)$ be a weighting function over iterations $t$ and $f(\cdot)$ be a function on which consistency loss is measured, then the consistency loss $L_U$ is computed in an unsupervised manner and consequently the total loss $L_{total}$ is given by a linear combination of the consistency loss and the supervised loss $L_S$ as follows: 
\begin{equation}
    \label{eq:consis_loss_diff}
    L_{U} = d(f(x), f(x'))
\end{equation}
\begin{equation}
    \label{eq:consis_loss}
    L_{total} = L_{S} + w(t) \cdot L_{U}.
\end{equation}
Some notable examples of consistency training include $\Pi$ model \cite{laine2016temporal}, virtual adversarial training \cite{miyato2018virtual} and Mean Teacher \cite{tarvainen2017mean}. The recent advances in this direction includes interpolation consistency training (ICT) \cite{verma2019interpolation} (its variants MixMatch \cite{berthelot2019mixmatch}, ReMixMatch \cite{Berthelot2020ReMixMatch}) and FixMatch \cite{sohn2020fixmatch}. 

ICT is a specific type of consistency regularization that encourages the prediction at an interpolation of unlabeled samples to be consistent with the interpolation of the predictions at those samples. FixMatch uses another form of consistency regularization, where the model's prediction on \quotes{weak augmentation} are encouraged to be consistent with the \quotes{strong augmentation}. For weak augmentation, FixMatch uses horizontal flipping, random translation and cropping, and for strong augmentation it uses Cutout \cite{devries2017cutout}, RandAugment \cite{cubuk2019randaugment} and CTAugment \cite{Berthelot2020ReMixMatch}.

\subsection{IR for Object Detection} 
Interpolation Regularization for Object Detection 
has recently been studied in \cite{zhang2019bag, bouabid2020mixup}.
They applied IR to object detection in a supervised manner, and they focused on the distribution and the mixed object region. However, they did not consider the relationship between a foreground object and the background (Our Type-II).
In this paper, different from the previous algorithms, we propose a method that applies IR to semi-supervised learning while considering the relationship between an object and the background.

\subsection{SSL for Object Detection}
Semi-Supervised Learning for Object Detection has recently been studied in \cite{jeong2019consistency} where CSD, the first consistency-regularization-based semi-supervised object detection method, was proposed.
They explored the consistency between the box predictions in the original and the horizontally flipped version. 
To prevent the ‘background’ class from dominating the consistency loss in Eq. (\ref{eq:consis_loss}), they proposed the Background Elimination (BE) method which excludes boxes with high background probability in the computation of the consistency loss. 
In this paper, we also utilize the BE using the class probability of each candidate box. Also, the proposed ISD is combined with CSD to produce the SOTA SSL object detection performance.

\begin{figure*}[t]
\begin{center}$
\centering
\begin{array}{c}
\includegraphics[width=0.8\linewidth]{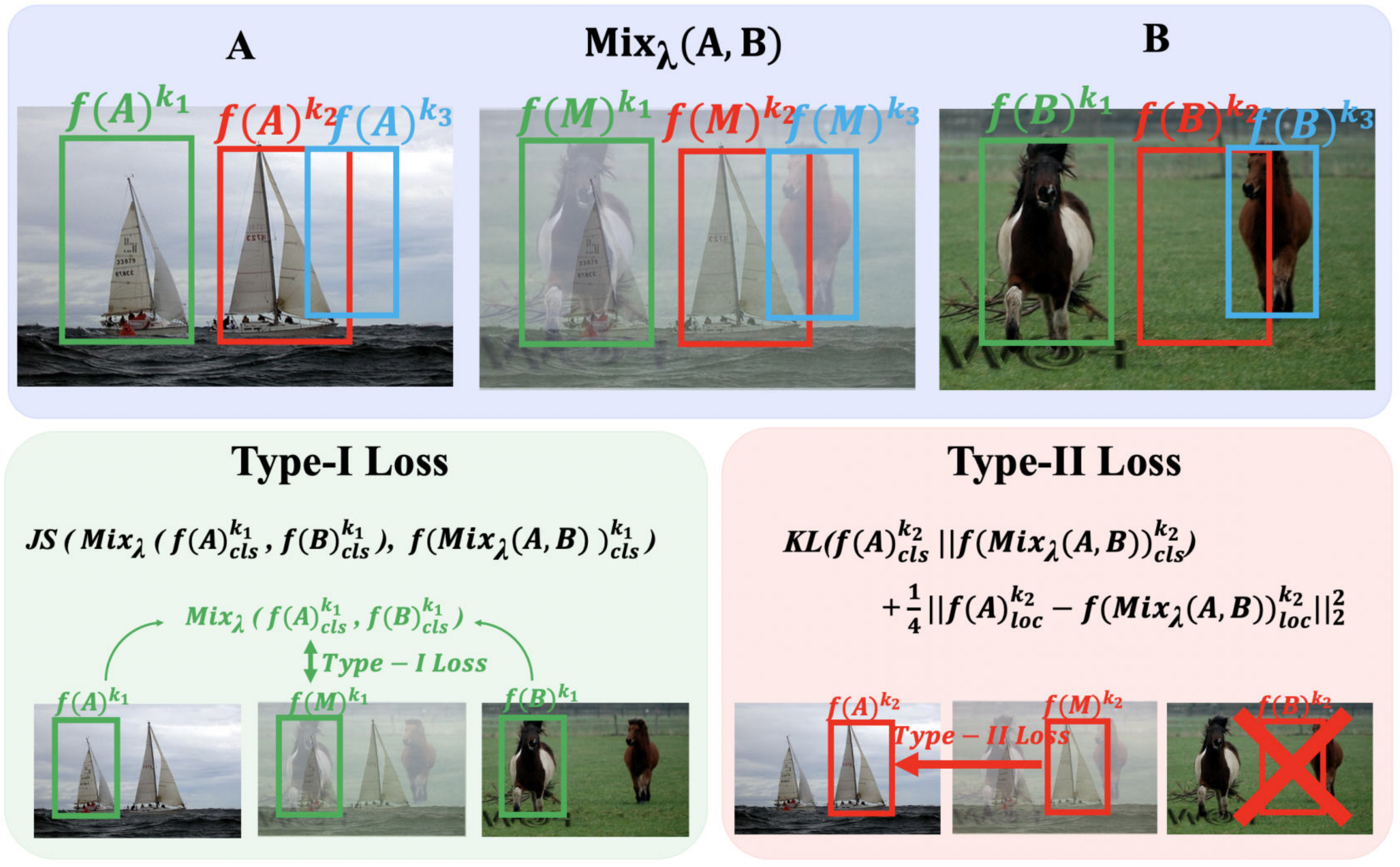}
\end{array}$
\end{center}
\caption{The proposed ISD loss for each type. $Mix_{\lambda}(a,b) = \lambda \cdot a + (1-\lambda) \cdot b$}
\label{method:isd}
\end{figure*}

\section{Method}
\label{sec:method}


We denote a horizontally flipped version of an image $A$ as $\hat{A}$, and the image created by random mixing, $\lambda \cdot A + (1-\lambda) \cdot B$, of two images $A$ and $B$ as $Mix_{\lambda}(A,B)$. Similar to Mixup, the mixing coefficient $\lambda$ is drawn from the  $Beta(\alpha,\alpha)$ distribution.
In our method, we use SSD \cite{liu2016ssd}, one of the most popular single-stage object detectors, as a baseline detector. 
In the training of SSD, we add the newly proposed interpolation-based consistency regularization loss in combination with the flip-based consistency regularization loss in \cite{jeong2019consistency} to enhance the performance. 
The network output of SSD $f^{p,r,c,d}$ is denoted as the output of the $p^{th}$ layer of the pyramid, $r^{th}$ row, $c^{th}$ column and $d^{th}$ default box, and ($p, r, c, d$) is expressed as $k$ for brevity.
Each $f^{k}$ is composed of $f^{k}_{cls}$ and $f^{k}_{loc}$ which are the softmax output vector and the localization offsets of the center and the size of the box, [$\Delta cx, \Delta cy, \Delta w, \Delta h$], at position $k$, respectively.
The mask $m(I)$, which is computed by $f_{cls}(I)$, is used in background elimination and interpolation type categorization for image $I$ and has the binary objectness value at each location $k$:
\begin{equation} \label{eq:eb}
m(I)^{k} = \begin{cases}
1, & \mbox{if } \ \text{argmax}(f_{cls}^{k}(I)) \neq background\\
0, & \mbox{otherwise}.
\end{cases}
\end{equation}

\subsection{Interpolation-based Semi-supervised learning for Object Detection (ISD)}

\subsubsection{Type categorization.}
We determine the type of a pair of patches by the background elimination method \cite{jeong2019consistency} that only extracts patches with a high objectness probability.
Then we apply different methods appropriate for each type of patches.
Eq.~(\ref{criteria_type}) is how we calculate each type of a mask.
The Type-I mask, $m_I$, is calculated by element-wise multiplication of $m(A)$ and $m(B)$.
In other words, it becomes 1 when both patches of $m(A)^{k}$ and $m(B)^{k}$ are 1, and otherwise it is 0.
On the other hand, the Type-II mask $m_{II}^A$ is calculated by element-wise multiplication of $m(A)$ and $\sim m(B)$, which means it is 1 when the patch in image $A$ has a high objectness score while the corresponding patch at the same location in image $B$ has a high background score, and vice versa for $m_{II}^B$.

\begin{equation} 
\begin{split}
\label{criteria_type}
\text{Type-I mask:} \quad &m_I =  m(A)  \otimes   m(B), \\
\text{Type-II(A) mask:} \quad &m_{II}^A = m(A) \otimes \sim m(B),  \\
\text{Type-II(B) mask:} \quad &m_{II}^B = \sim m(A) \otimes m(B).  
\end{split}
\end{equation}

\subsubsection{Type I loss}
%
When the patches in image $A$ and $B$ are all likely to be objects (Type-I), we define a Type-I loss inspired by the ICT loss \cite{verma2019interpolation}.
Note that there are two differences compared to the conventional ICT.
First, we used \textit{Jensen-Shannon divergence} (JSD) as the consistency regularization loss (function $d(.,.)$ in Eq. (\ref{eq:consis_loss})).
Second, we use the same network to feed-forward inputs like CSD, distinct from ICT which uses different networks for mixed and original inputs using MeanTeacher \cite{tarvainen2017mean}.
Eq.~(\ref{loss_type1}) shows the loss function of Type-I, which is the distance between the mixed output of $f(A)_{cls}^k$ and $f(B)_{cls}^k$ and the output of the mixed image of $A$ and $B$, $f(Mix_{\lambda}(A, B))_{cls}^k$. 
\begin{equation} 
\label{loss_type1}
\begin{split}
 l_{I} = JS ( Mix_{\lambda}(f(A)_{cls}^k, f(B)_{cls}^k) || f(Mix_{\lambda}(A, B))_{cls}^k )
\end{split}
\end{equation}
The overall Type-I loss $\mathcal{L}_I$ is the average of patches whose Type-I mask is 1, i.e.
$
\mathcal{L}_{I} 
= \mathbb{E}_{\mathbb{I}{\{m_I = 1\}}} [l_{I}] .
$ 
Here, $\mathbb{E}$ and $\mathbb{I}$ are the expectation and the indicator function, respectively.

\subsubsection{Type II loss}
As shown in Fig.~\ref{method:isd}, in Type II, one patch has a high probability of foreground, while the other has a high probability of background. 
In this case, instead of using the Type I loss described above, we train the network to have similar predictions on the mixed patch and the patch with a high probability of foreground.
This kind of loss can be interpreted as a form of FixMatch loss \cite{sohn2020fixmatch} which encourages  consistency between the predictions on the strong augmentation and the weak augmentation of an input. 
More specifically, in our case, the mixed patch is considered as a strong augmentation while the patch with a high foreground probability acts as no-augmentation. Note that, for classification, FixMatch is trained with targets by creating pseudo-labels of samples that exceed a threshold, whereas we do not need to set a specific threshold and the target is set according to the output distribution of no-augmentation patch.

We set $f(A)$ or $f(B)$ as a target, and train the mixed output ($f(Mix_{\lambda}(A,B))$) to be close to $f(A)$ or $f(B)$.
In doing so, Kullback-Leibler (KL) divergence and L2 loss are used as the classification and localization losses, respectively as follows: 
\begin{equation} 
\label{typeII_loss_c}
l_{II\_cls}^A = KL (f(A)_{cls}^k|| f(Mix_{\lambda}(A,B))_{cls}^k)
\end{equation}
\begin{equation} 
\label{typeII_loss_l}
l_{II\_loc}^A = \frac{1}{4} \lVert f(A)_{loc}^k - f(Mix_{\lambda}(A,B))_{loc}^k \lVert _{2}^{2}.
\end{equation}
%
The overall Type-II loss when patch $A$ is foreground, $\mathcal{L}_{II}^A$, is calculated as the average of the sum of two individual losses as
$\mathcal{L}_{II}^A 
= \mathbb{E}_{\mathbb{I}{\{m_{II}^A = 1\}} 
} [l_{II\_cls}^A + l_{II\_loc}^A]$.
Likewise, $\mathcal{L}_{II}^B$ is also calculated by applying the above loss, and the total loss of Type-II is calculated as
$\mathcal{L}_{II} 
= \mathcal{L}_{II}^A  + \mathcal{L}_{II}^B 
$.

Finally, the overall ISD loss is computed by Type-I loss ($\mathcal{L}_{I}$) and Type-II loss ($\mathcal{L}_{II}$) as follows: 
\begin{equation} 
\label{Total_isd_loss}
\begin{split}
 \mathcal{L}_{ISD} = \gamma _{1} \cdot \mathcal{L}_{I} + \gamma_{2} \cdot \mathcal{L}_{II}.
\end{split}
\end{equation}
Here, $\gamma_1$ and $\gamma_2$ are set appropriately to balance both loss terms.

\begin{algorithm}[t]
\caption{Training procedure of the proposed ISD }
\label{algorithm:isd}
\textbf{Require}: $\mathcal{D_{L}}, \mathcal{D_{U}}$: labeled and unlabeled datasets
\\
\textbf{Require}: $w(t)$: weight scheduling function
\\\textbf{Require}: $f(\cdot)$: trainable object detection model 
\\\textbf{Require}: $h(\cdot)$: horizontal flip function
\\\textbf{Require}: $m(\mathcal{\cdot})$: objectness masks

\begin{algorithmic}[1]
\FOR {each $t \in [1$, max\_iterations$]$}
\STATE \textit{\textbf{Data Preparation}}
\STATE \quad $\mathcal{A} \gets \mathcal{D_L} \cup \mathcal{D_U}$, $\hat{\mathcal{A}} \gets h(\mathcal{A})$ 
\STATE \quad $\mathcal{B} \gets shuffle(\hat{\mathcal{A}})$ 
\STATE \quad $\mathcal{C} \gets Mix_{\lambda}(\mathcal{A}, \mathcal{B})$ 
\\
\STATE \textit{\textbf{Compute the outputs}}
\STATE \quad $f(\mathcal{A}), f(\hat{\mathcal{A}}), f(\mathcal{C}) $ 
\STATE \quad $f(\mathcal{B}) \gets shuffle(f(\hat{\mathcal{A}}))$
\STATE \textit{\textbf{Compute the objectness mask}}
\STATE \quad $m_{\mathcal{A}} \gets f(\mathcal{A}),\quad m_{\mathcal{B}} \gets f(\mathcal{B}) $ \qquad \quad $(Eq.~\ref{eq:eb})$
\STATE \textit{\textbf{Compute the supervised \& CSD losses}}
\STATE \quad $ \mathcal{L}_{S} \gets f(A \in \mathcal{D_{L}} \cap \mathcal{A})$
\STATE \quad $ \mathcal{L}_{CSD} \gets f(A \in \mathcal{D_{U}} \cap \mathcal{A}), f(\hat{A}), m_{\mathcal{A}}$
\STATE \textit{\textbf{Compute the ISD loss using the type mask (Eq.~\ref{criteria_type}) }}
\STATE \quad $ \mathcal{L}_{I} \gets  \mathbb{E}_{\mathbb{I}{\{m_I = 1\}}} [l_{I}] \quad \quad \quad \quad \quad \quad \quad(Eq.~\ref{loss_type1}) $ 
\STATE \quad $ \mathcal{L}^{A}_{II} \gets  \mathbb{E}_{\mathbb{I}{\{m^{A}_{II} = 1\}}} [l^{A}_{II\_cls} + l^{A}_{II\_loc}] \quad (Eq.~\ref{typeII_loss_c},\ref{typeII_loss_l})$
\STATE \quad $ \mathcal{L}^{B}_{II} \gets  \mathbb{E}_{\mathbb{I}{\{m^{B}_{II} = 1\}}} [l^{B}_{II\_cls} + l^{B}_{II\_loc}]$
\STATE \quad $ \mathcal{L}_{II} \gets  \mathcal{L}^{A}_{II} + \mathcal{L}^{B}_{II}$
\STATE \quad $ \mathcal{L}_{ISD} \gets  \lambda_{1} \cdot \mathcal{L}_{I} + \lambda_{2} \cdot \mathcal{L}_{II}$

\STATE \textit{\textbf{Compute the total loss}}
\STATE \quad $ \mathcal{L}_{Total} \gets \mathcal{L_{S}} + w(t)\cdot(\mathcal{L}_{CSD} + \mathcal{L}_{ISD}) $
\STATE \textit{\textbf{Update $f(\cdot)$ using $\mathcal{L}_{Total}$}}
\ENDFOR
\end{algorithmic}
\end{algorithm}

\begin{table*}[ht]
\begin{center}
\caption{Detection results for PASCAL VOC2007 test set under the supervised and the semi-supervised training setting.
$L_{cls}$ and $L_{loc}$ are the consistency classification and localization loss with BE (Eq.~\ref{eq:eb}) in CSD.
The following experiments use VOC07 (labeled) and VOC12 (unlabeled) data.
\textcolor{blue}{Blue} and \textcolor{red}{Red} are represented as the Baseline score and Best score, respectively. The numbers in the parentheses are the performance increments compared with the baseline.}
\vspace{0.2cm}
\label{tab:expvoc}
\begin{tabular}{|l||c|c||c|}
\hline
Semi-Supervised Loss & Labeled data & Unlabeled data & mAP (\%)\\
\hline
\hline
\multicolumn{4}{|c|}{Supervised Learning -- Trained only with labeled data}\\
\hline
None (Supervised Learning) & VOC07   & - & \textcolor{blue}{70.2}  \\
\cite{liu2016ssd, jeong2019consistency} & VOC07 + VOC12 & - & 77.2 \\
\hline
CSD \cite{jeong2019consistency} & \multirow{3}{*}{VOC07} & - & 69.3 \\
Ours (ISD only)  &  & - & 72.3 \\
Ours (ISD + CSD)  &  & - & 73.1 \\
\hline
\hline
\multicolumn{4}{|c|}{Semi-Supervised Learning}\\
\hline
CSD \cite{jeong2019consistency} ($L_{cls}$) & \multirow{3}{*}{VOC07} & \multirow{3}{*}{VOC12} & 71.7 (1.5)\\
CSD \cite{jeong2019consistency} ($L_{loc}$) &  &  & 71.9 (1.7)\\
CSD \cite{jeong2019consistency} ($L_{cls}$ + $L_{loc}$) &  &  & 72.3  (2.1)\\
\hline
Ours (ISD (Type-I only)) & \multirow{4}{*}{VOC07} & \multirow{4}{*}{VOC12} & 71.9  (1.7)\\
Ours (ISD (Type-II only)) &  &  & 73.8 (3.6)\\
Ours (ISD (Type-I,II)) &  &  & 74.1 (3.9)\\
Ours (CSD + ISD (Type-I,II)) &  &  &  \textcolor{red}{74.4 (4.2)} \\

\hline
\end{tabular}
\end{center}
\end{table*}

\begin{figure}[t]
\begin{center}$
\centering
\begin{array}{c}
\includegraphics[width=8cm]{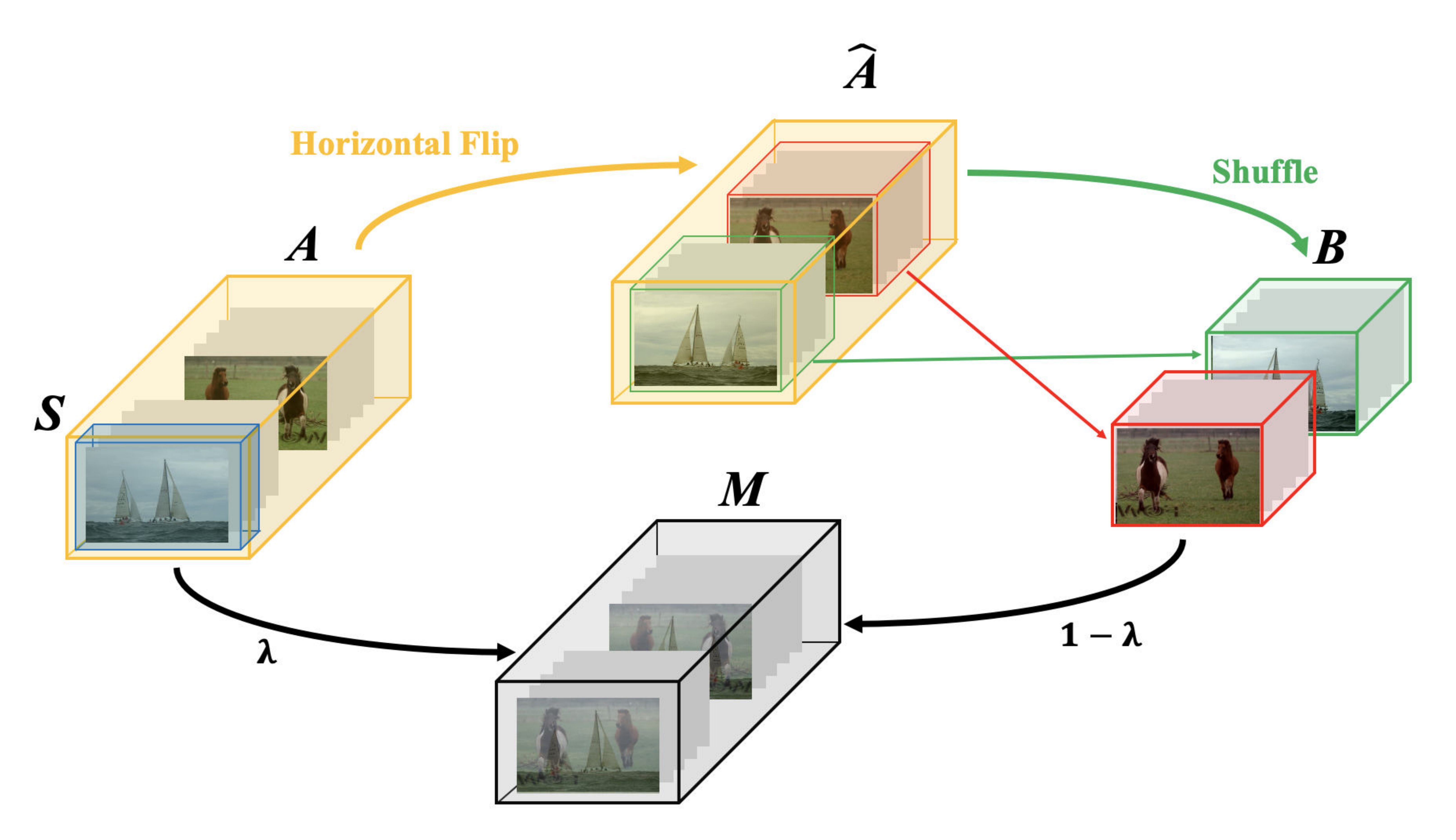}
\end{array}$
\end{center}
\caption{Combination of ISD with CSD. The original images ($\mathcal{A}$) are flipped ($\hat{\mathcal{A}}$) and the mixed images ($\mathcal{M}$) are obtained by combining the two. First, the order of flipped images are changed by shuffling  ($\mathcal{B} = \text{shuffle}(\hat{\mathcal{A}}$)), then $\mathcal{A}$ and $\mathcal{B}$ are mixed ($\mathcal{M} = {Mix}_\lambda (\mathcal{A}, \mathcal{B})$). CSD loss is calculated between $\mathcal{A}$ and $\hat{\mathcal{A}}$ and ISD loss is computed between $\mathcal{M}$ and ($\mathcal{A}$ and/or $\mathcal{B}$). In the original set ($\mathcal{A}$), the blue box ($\mathcal{S}$) is labeled, to which the supervised loss is applied.}
\label{method:combination}
\end{figure}

\subsection{Combination of ISD with CSD}

For ISD training, three sets of image batches, $\mathcal{A}$, $\mathcal{B}$, and $\mathcal{M} = Mix_{\lambda}(\mathcal{A},\mathcal{B})$ should be inferred by the network.
For efficient training, we set $\mathcal{B}$ as the horizontally flipped version of ${\mathcal{A}}$, i.e, $\hat{\mathcal{A}} = {flip}(\mathcal{A})$, as shown in Fig.~\ref{method:combination}. We calculated the CSD loss with those two batches.
However, the mixed image $Mix_{\lambda}(A,\hat{A})$ of $A \in \mathcal{A}$ and its horizontal flipped version $\hat{A} \in \hat{\mathcal{A}}$ would have similar backgrounds and predict the same class in the center of the image.
Therefore, as shown in Fig.~\ref{method:combination}, we make the mixed images by combining the original batch ($\mathcal{A}$) with the half-shuffled flipped batch ($\mathcal{B} = {shuffle}(\hat{\mathcal{A}}$)).
The total loss consists of supervised loss ($\mathcal{L}_{S}$), CSD loss ($\mathcal{L}_{CSD}$), and ISD loss ($\mathcal{L}_{ISD}$) as follows:
\begin{equation} 
\label{Total_loss}
\begin{split}
 \mathcal{L}_{Total} = \mathcal{L}_{S} + w(t) \cdot [\mathcal{L}_{CSD} + \mathcal{L}_{ISD}],
\end{split}
\end{equation}
where $w(t)$ is a weight scheduling function. 
The overall process of the proposed semi-supervised learning is described in Algorithm ~\ref{algorithm:isd}

\section{Experiments}
\label{sec:experiments}

\subsection{Experimental Settings}
Our experiments are based on pytorch.
We have used a third-party code for SSD\footnote{\url{https://github.com/amdegroot/ssd.pytorch}} and an official code for CSD\footnote{\url{https://github.com/soo89/CSD-SSD}}.
We experimented on the PASCAL VOC dataset and MS COCO dataset with SSD300 model. VGG-16 pre-trained model is used as our backbone network.
PASCAL VOC \cite{everingham2010pascal} and MS COCO \cite{lin2014microsoft} data consist of 20 and 80 classes, respectively.
For VOC dataset, we followed the settings from the conventional semi-supervised learning methods for object detection.
Similar to \cite{wang2018towards, jeong2019consistency}, we trained our model with PASCAL VOC07 \textit{trainval} (5k images) dataset as labeled data and PASCAL VOC12 \textit{trainval} (12k images) as unlabeled data. \nj{Then, we} tested with PASCAL VOC07 test dataset.
For MS COCO dataset, we divided the MS COCO 2014 dataset into the existing categorized Train2014 (83k images) and Val2014-35k (35k images) dataset because minor classes may not be in the labeled dataset with random sampling.
We trained our model with Val 35k dataset as labeled data and Train 83k as unlabeled data. Then, we tested with MS COCO test-dev dataset.


We sample the mixing parameter $\lambda$ from $Beta$($\alpha$, $\alpha$) at every iteration. The parameters are set to  $(\gamma_{1}$, $\gamma_{2})$ = (0.1, 1) in Eq.~(\ref{Total_isd_loss}) and $\alpha$ = 100 in \nj{the} beta distribution.
Other learning parameters such as \nj{the} learning rate and \nj{the} batch size are the same as \cite{jeong2019consistency}.
In our experiment, we report the mean and standard deviation of three runs.

\subsection{PASCAL VOC}
\label{exp::pascal}
\subsubsection{Supervised Learning}
\label{exp:sup}
We start by examining the effect of ISD on SSD in the supervised training setting, i.e, the proposed losses in \ref{Total_loss} are applied to labeled data. The results are presented in Table~\ref{tab:expvoc}.
In the first row block, SSD (base) trained with VOC 07 (\textit{trainval}) data shows 70.2 mAP performance, while that of SSD (CSD) decreases to 69.3 mAP, which shows a clear side effect of over-regularization.
On the other hand, SSD300 (ISD) and SSD (ISD + CSD) show 2.1\% and 2.9\% improvements in accuracy compared to SSD (base), respectively.
This shows that combining ISD with a strong CSD regularizer stabilizes the training, making the network more robust.

\begin{table*}[t]
\begin{center}
\caption{Detection results for MS COCO test-dev set.
The following experiments use Val35k (labeled) and Train80k (unlabeled) data. The numbers in the parentheses are the performance improvements from the baseline model (SSD trained on Val35k). All experiments are tested by ourselves.
}
\label{tab:expcoco}
{
\begin{tabular}{|l||c|c||ccc|}
\hline
\multirow{2}{*}{Method} & Labeled  & Unlabeled  &  \multicolumn{3}{c|}{Avg. Precision, IoU:}\\
\cline{4-6}
& data & data & 0.5:0.95 & 0.5 & 0.75\\
\hline
\hline
\multirow{2}{*}{SSD \cite{liu2016ssd}} & Val35k & - & 18.8 & 34.8 & 18.6 \\
 & Val35k + Train80k (trainval35k) &- & 23.9 & 40.8 & 24.7 \\
 \hline
CSD \cite{jeong2019consistency} & \multirow{2}{*}{Val 35k} & \multirow{2}{*}{Train 80k} & 19.8 (1.0) & 35.8 (1.0) & 19.8 (1.2) \\
Ours (CSD + ISD) & & & \textcolor{red}{21.0 (2.2)} & \textcolor{red}{37.7 (2.9)} & \textcolor{red}{21.1 (2.5)} \\

\hline
\end{tabular}
}
\vspace{-4mm}
\end{center}
\end{table*}

\subsubsection{Semi-Supervised Learning}
\label{exp:semi}

We evaluate the performance of ISD in the SSL setting.
As shown in Table~\ref{tab:expvoc}, the performance of the SSD model trained only with VOC07 labeled data is 70.2\%.
Type-I and Type-II show 1.7\% and 3.6\% of enhancement, respectively.
The Type-I consists of only classification loss, and it shows better result than the score of only classification loss in CSD.
Type-II shows much better performance than CSD and jointly using both Type-I and Type-II losses shows 3.9\% of enhancement.
In addition, when CSD and ISD are combined, it shows even greater performance improvement.
This demonstrates the effectiveness of our approach in the SSL setting.
%
Moreover, ISD+CSD with VOC07 labeled data and VOC12 unlabeled data on SSD (Table~\ref{tab:expvoc}, last row) shows 1.3\% performance improvement in comparison to the fully supervised setting with VOC07 labeled data on SSD (Tabel~\ref{tab:expvoc}, row 7).
This explains that the combined loss of ISD+CSD not only on labeled data, but also on unlabeled data contributes to better performance. 
The results shown in Table~\ref{tab:expvoc} demonstrate that our ISD+CSD approach outperforms the baseline CSD-only approach by a significant margin.

\subsection{MSCOCO}
\label{exp:mscoco}

Table~\ref{tab:expcoco} shows the results of experiments on the MSCOCO dataset.
The supervised performances of SSD using Val35k and Trainval35k show 18.8 mAP and 23.9 mAP, respectively.
CSD with Val35k labeled data and Train80k unlabeled data on SSD shows 1.0\% of enhancement. 
Our proposed algorithm (CSD+ISD) shows 2.1\% performance improvement in the same experimental setting for COCO dataset.


\section{Discussion}
\label{sec:discussions}



\subsection{Ablation studies for Type-I and Type-II losses}
\label{dis:type}

\begin{figure*}[t]
\begin{center}$
\centering
\begin{tabular}{p{0.1cm} c c  c  c  c} 

\rotatebox{90}{\qquad \quad \textbf{SSD}}& %
\includegraphics[width=3.2cm, height=2.5cm]{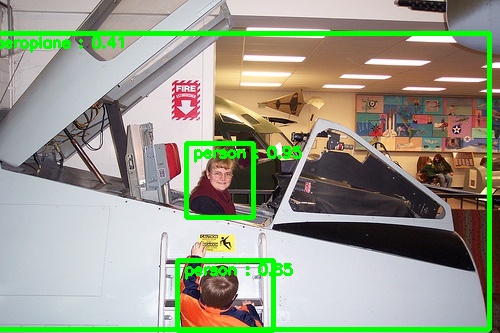} & \hspace{-0.4cm}
\includegraphics[width=3.2cm, height=2.5cm]{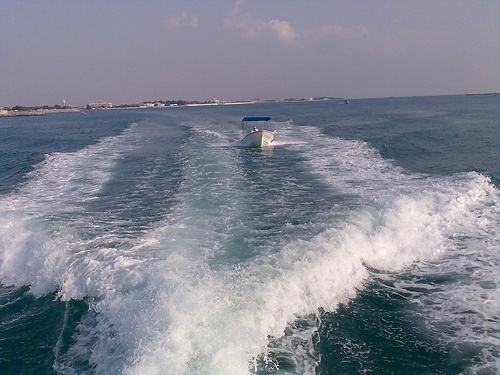} & \hspace{-0.4cm}
\includegraphics[width=3.2cm, height=2.5cm]{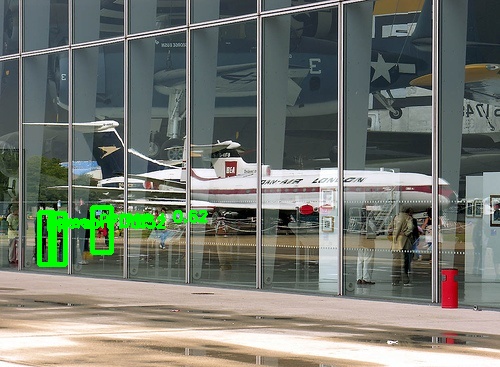} &  \hspace{-0.4cm}
\includegraphics[width=3.2cm, height=2.5cm]{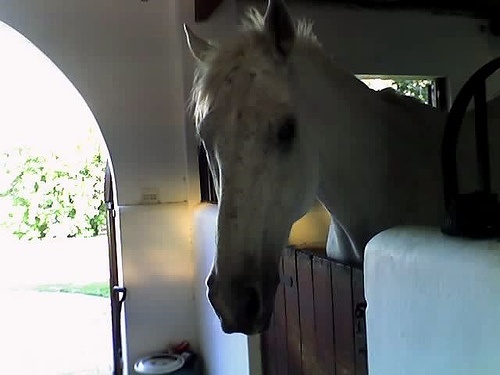} & \hspace{-0.4cm}
\includegraphics[width=3.2cm, height=2.5cm]{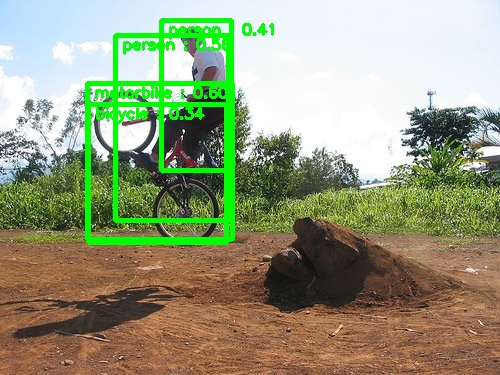} \\
\rotatebox{90}{\qquad \quad \textbf{CSD}}& %
\includegraphics[width=3.2cm, height=2.5cm]{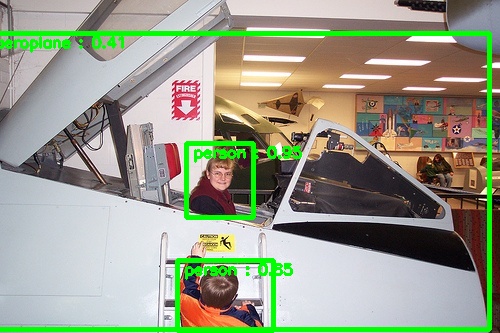} & \hspace{-0.4cm}
\includegraphics[width=3.2cm, height=2.5cm]{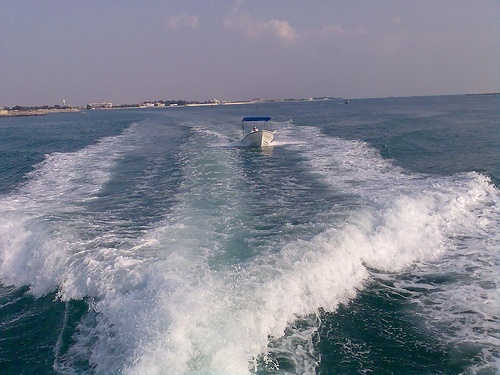} & \hspace{-0.4cm}
\includegraphics[width=3.2cm, height=2.5cm]{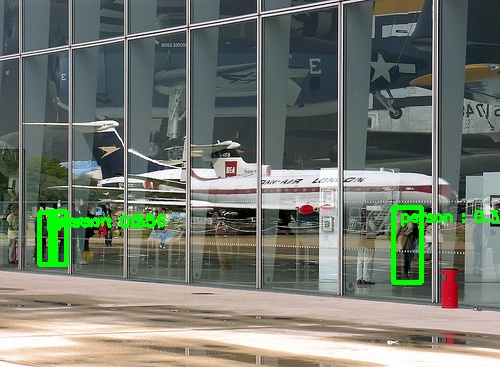} & \hspace{-0.4cm}
\includegraphics[width=3.2cm, height=2.5cm]{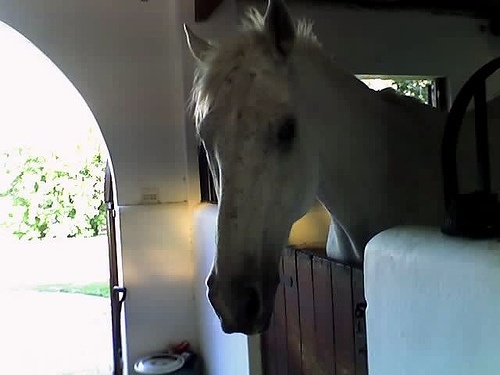} & \hspace{-0.4cm}
\includegraphics[width=3.2cm, height=2.5cm]{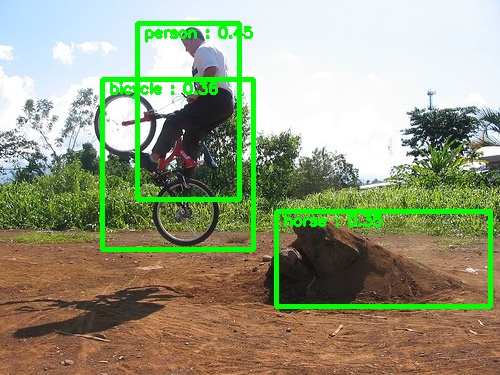} \\

\rotatebox{90}{\qquad \textbf{CSD+ISD}}& %
\includegraphics[width=3.2cm, height=2.5cm]{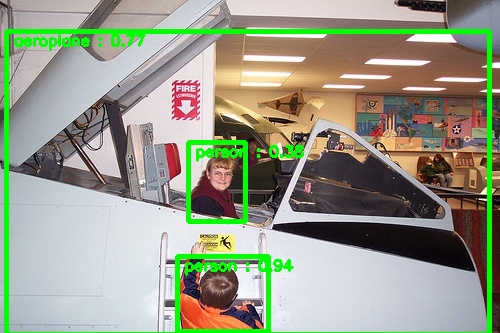} & \hspace{-0.4cm}
\includegraphics[width=3.2cm, height=2.5cm]{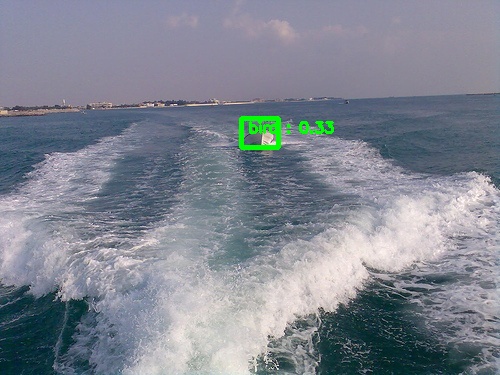} & \hspace{-0.4cm}
\includegraphics[width=3.2cm, height=2.5cm]{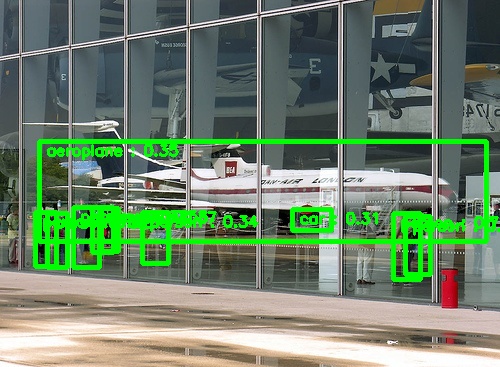} & \hspace{-0.4cm}
\includegraphics[width=3.2cm, height=2.5cm]{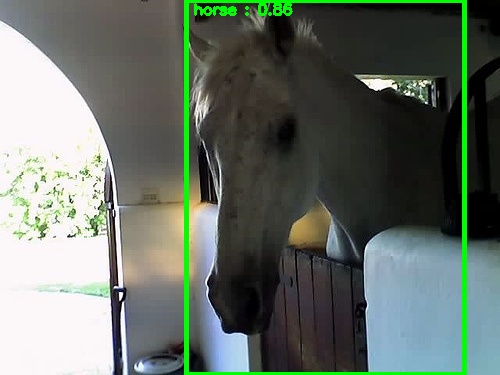} & \hspace{-0.4cm}
\includegraphics[width=3.2cm, height=2.5cm]{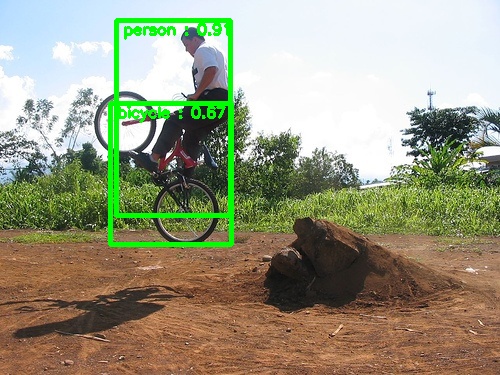} \\

\end{tabular}$
\end{center}
\caption{Qualitative results for the PASCAL VOC2007 test set using supervised SSD, semi-supervised CSD and CSD+ISD models in table~\ref{tab:expvoc}. The first, middle, and last rows are the resulting images of the SSD, CSD, and CSD+ISD models, respectively. A score threshold of 0.3 is used to display these images. The images from the second column to the fourth column are the result when the image of the object is similar to the background or there is distortion. Our proposed algorithm shows that it works robustly in this situation. The results of the last column show that ISD does not detect all samples that look like objects.}
\label{exp:example_image}
\end{figure*}

We experiment to verify the performance of the two types of loss we proposed in Table~\ref{tab:expvoc}.
Each loss shows a significant performance improvement compared to the supervised learning.
Furthermore, we report the combination of CSD and the different types of ISD losses in Table~\ref{exp:beta}.
In the table, for all the cases, the Type-II loss performed better than of Type-I loss.
There are three reasons for this results.
First, the numbers of Type-I and Type-II samples are different.
With a trained model, the number of Type-II samples was 5 times that of Type-I samples, which indicates that the influence of Type-I loss is relatively small.
Second, Type-I only considers the classification loss while Type-II uses the localization loss as well.
Because the two objects in Type-I have different bounding boxes, the boundary of their mixed patch is not equal to the interpolation of their bounding boxes.
Therefore, the localization loss cannot be applied in Type-I cases.
Third, two objects that are mixed may not be aligned well.
More research is needed for the alignment in Interpolation Regularization, which remains as a future work.

In Table~\ref{exp:typeII_ablation}, we analyzed the effect of the classification and the localization loss in Type-II when $\alpha$ is 100.
The classification loss on Type-II samples showed more remarkable performance improvement than the localization loss, and by combining them, we can obtain better performance.

\begin{table}[t]
\caption{
Ablation study for $\alpha$ and each type in VOC07(L) + VOC12(U) training dataset and VOC07 testing dataset.
The row represents the $\alpha$ of the beta distribution, and the column represents each type.
All the experiments in this table are performed by adding each loss to the CSD.
}
\centering
\label{exp:beta}
{
\begin{tabular}{|c||c|c|c|}
\hline
$\beta$($\alpha$, $\alpha$) & 
\multicolumn{3}{c|}{SSD300 + ISD Method (mAP (\%))} \\
\cline{1-1}\cline{2-4}
$\alpha$ &  Type-I & Type-II & Type-I + Type-II \\
\hline
\hline
1 & 72.3 & 72.8 & 72.9\\
10 & \textbf{72.4} & 73.8 & 74.0\\
100 & \textbf{72.4} & \textbf{74.2} & \textbf{74.4}\\
1000 & 72.2 & \textbf{74.2} & 74.3\\
\hline
\end{tabular}
}
\end{table}

\begin{table}[t]
\caption{
Ablation study of Type-II losses on PASCAL VOC2007 test set.
All the experiments in this table are performed by adding each loss to the CSD.
($\alpha$ is 100).
}  
\centering
\label{exp:typeII_ablation}
\begin{tabular}[r]{|l|c|}
\hline
\scriptsize{VOC07(L)+VOC12(U)} & mAP (\%)\\
\hline
\hline
Type-II (cls) & 74.0 \\
Type-II (loc) & 73.1\\
Type-II (cls + loc) & \textbf{74.2}\\
\hline
\end{tabular}
\end{table}

\subsection{Beta distribution}
\label{dis:beta}

In ISD, the mixing coefficient $\lambda$ is sampled from the $Beta(\alpha, \alpha)$ distribution. 
Table~\ref{exp:beta} shows the performance of ISD using various values of $\alpha$ across different types of ISD losses. 
We observe that a large range of $\alpha$ gives improved performance in comparison to the baseline (CSD with 72.3\% mAP). 
In general, we recommend to set $\alpha$ to a  sufficiently large value. 
The reason for choosing relatively large $\alpha$ is as follows:
With a smaller values of $\alpha$ (e.g. $\alpha < 1$), $\lambda$ will be close to either 0 or 1 with high probability, thus most of the mixed images will be closer to either of the original images being mixed. In this case, the mixed image $M$ will be extremely weak (for one image) or strong (for the other) augmentation resulting in lowered performance with high variance. 
In contrast, increasing the values of $\alpha$ increases the probability of $\lambda$ being closer to 0.5, which provides an appropriate level of regularization. 
Note that if the value of $\alpha$ is too large, 
$\lambda$ will be concentrated too much around 0.5 
and all the augmented samples will be too different from the original images resulting in degraded performance with high variance at test time.  


\subsection{Training model size}
\label{dis:size}

For ISD training, image batches are inferred by the network three times over conventional SSD. 
Also, due to the calculation of additional losses, it requires more than three times the conventional SSD memory. 
We used Nvidia 1080Ti GPU, and we assigned 4 GPUs for SSD model with ISD training. 
With fewer GPUs, our implementation was not trainable because of limited memory budget.
However, at testing, it has the same network size and inference time as the base network and can improve the performance.

\subsection{Object detector}
\label{dis:detector}

In this paper, we have used the SSD model among various single stage detectors.
In the case of other detectors, algorithm-specific modifications should be made to successfully apply interpolation regularization, while the basic idea of separating Type-I and Type-II samples and applying a different loss for each case is still valid.
In the case of a Two-Stage detector, generally, Region of Interest (RoI) is obtained by Region Proposal Network (RPN) and classification of that location is performed for object detection.
Since the RoIs of $A$, $\hat{A}$, $B$, and $Mix_{\lambda}(A,B)$ are all different, in order to apply our algorithm, one of RoIs should be applied to other images for one-to-one correspondence.
If the RoI of $A$ is applied to other images, Type-II loss between $B$ and $Mix_{\lambda}(A,B)$ cannot be obtained, and if each RoI of \textit{A}, \textit{B}, $Mix_{\lambda}(A,B)$ is applied individually to other images, a lot of computation will be required.
Thus how to apply interpolation-based regularizer for Two-stage detectors is an interesting avenue for further research.

\section{Conclusion}
\label{sec:conclusion}

In this paper, we have proposed ISD, a simple and efficient Interpolation-based semi-supervised learning algorithm for object detection using single-stage detectors. We started by investigating the challenges that occur when the Interpolation Regularization methods for the classification task are applied directly to an object detection task, and have addressed these challenges by proposing different types of interpolation-based loss functions. Our method shows significant improvement in both semi-supervised and supervised object detection tasks over the previous methods, compared over the same dataset and the same architecture settings. We further demonstrate that combining ISD with the previous method of CSD can further improve the performance. We leave the exploration of Interpolation Regularization for Two-stage detectors as a future work.

\newpage
{\small
\bibliographystyle{ieee_fullname}
\bibliography{egbib}
}

\end{document}